\documentclass[preprint]{elsarticle}

\journal{Journal of \LaTeX\ Templates}

\biboptions{authoryear}

\usepackage{amsmath}
\usepackage{graphicx}
\usepackage{times}
\usepackage[english]{babel}
\usepackage{epstopdf}
\usepackage{algorithm}
\usepackage{algorithmic}
\usepackage{booktabs}
\usepackage{verbatim}
\usepackage{amsfonts}
\usepackage{subfigure}
\DeclareMathOperator*{\argmin}{arg\,min}

\usepackage{amssymb}
\usepackage[font=small,labelfont=bf]{caption}

\begin{document}
	
	\begin{frontmatter}
		
		\title{Bottleneck Supervised U-Net for Pixel-wise Liver and Tumor Segmentation }
		
		\author[rvt]{Song LI\corref{cor1}}
		\cortext[cor1]{Corresponding author. This work was partly done while the author was at Tencent.}
		\ead{sli228-c@my.cityu.edu.hk}
		
		\author[rvt]{Geoffrey K. F. Tso}
		\ead{msgtso@cityu.edu.hk}
		
		\address[rvt]{College of Business, City University of Hong Kong}

		\begin{abstract}
			In this paper, we propose a bottleneck supervised (BS) U-Net model for liver and tumor segmentation. Our main contributions are: first, we propose a variation of the original U-Net that incorporates dense modules, inception modules and dilated convolution in the encoding path; second, we propose a bottleneck supervised (BS) U-Net that contains an encoding U-Net and a segmentation U-Net. To train the BS U-Net, the encoding U-Net is first trained to get encodings of the label maps that contain the anatomical information (shape and location). Subsequently, this information is used to guide the training of the segmentation U-Net so as to reserve the anatomical features of the target objects. More specifically, the loss function for segmentation U-Net is set to be the weighted average of the dice loss and the MSE loss between the encodings and the bottleneck feature vectors. The model is applied to a public liver and tumor CT scan dataset. Experimental results show that besides achieving excellent overall segmentation performance, BS U-Net also works great in controlling shape distortion, reducing false positive and false negative cases. 
			
		\end{abstract}
		
		
	\end{frontmatter}
	
	
	\section{Introduction}
	Convolutional Neural Network (CNN) and Deep Convolutional Neural Network (DCNN) have received tremendous attention in recent years because it can learn very complicated features from the training dataset. Neural network can trace its history back to \cite{mcculloch1943logical}, in which the first mathematical model of neural network was proposed. After experiencing years of ``AI winter'' around 1980s, it has regained popularity in recent years because of the increasing data volume and the development of the required hardware (such as multi-core CPU and GPU). There are many deep learning applications, such as object detection (\citeauthor{ren2017faster} 2017; \citeauthor{ redmon2017yolo9000} 2017), style transfer(\citeauthor{luan2017deep} 2017; \citeauthor{gatys2016image} 2016) and image semantic segmentation (\citeauthor{chen2018deeplab} 2018; \citeauthor{long2015fully} 2015; \citeauthor{ronneberger2015u} 2015). In this paper, we focus on its application in liver and liver tumor segmentation in CT images.
	
	Automatic segmentation of medical images is a long-standing issue and is still being addressed by researchers. Medical imaging is a technique for visualizing the interior of a body for clinical analysis and medical intervention. In recent decades, medical imaging techniques such as X-ray, CT, MRI and ultrasound have been developed, based on which radiologists and physicians can make fast and accurate diagnosis. Even though accuracy of manual diagnosis has been greatly improved by these techniques, it still depends highly on radiologists' expertise. Therefore, misdiagnosis happens often and this always leads to mortal consequences. An automatic segmentation method is urgently needed for clinical practice. Fortunately, this need is most likely to be fulfilled by the fast developing computer vision techniques. 
	
	Computer vision technologies help computers gain high-level understanding of digital images or videos so as to relieve manual work and reduce errors of judgement. In general, they lie in the following two categories: supervised methods and unsupervised methods. The algorithmic foundations of traditional computer vision were formed in 1970s, including non-polyhedral, polyhedral modeling, motion estimation, edges extraction and so on. In recent years, computer vision is becoming increasingly popular thanks to the development of convolutional neural network (CNN). Nowadays, computer vision related tasks (such as image classification, semantic segmentation, object detection, deep reinforcement learning and so on) have been widely studied in academia as well as in industry. 
	
	
	Taking advantage of the computer vision techniques, researchers are making computers automatically analyze the images generated from medical devices, trying to achieve results comparable to or even better than physicians.
	
	\subsection{Related Work}
	Automatic medical image segmentation methods have been widely studied in the literature. These researches mainly fall into two categories: supervised methods and unsupervised methods.
	
	Unsupervised methods learn patterns of the data without referring to the ground truth. In \cite{stawiaski2008interactive}, a sequence of unsupervised techniques were used to conduct liver tumor segmentation. More specifically, a sub-volume is manually defined at first, then a region adjacency graph is extracted using watershed segmentation. Later, the liver boundaries are extracted using the minimal surfaces technique. Finally, the tumors are segmented using MAP estimation of a Markov random field (MRF). \cite{li2013likelihood} used a complex level set model to semi-automatically segment hepatic tumors from contrast-enhanced clinical CT images. \cite{li2012new} also used a level set model to integrate image gradient, region competition and prior information for CT liver tumor segmentation. \cite{lipkova2017automated} used a phase separation approach. They assumed the healthy phase and lesion phase images are polluted by noises, and then removed the noises and separated the mixture using Cahn-Hilliard equation. In \cite{das2016kernelized}, the adaptive threshold, morphological processing, and kernel fuzzy C-means (KFCM) clustering algorithm were used to visualize and measure the tumor areas from abdominal CT images. One of the advantages of unsupervised methods is that they have more generalizability since they are not learned from a certain population. However, the performance of unsupervised methods can be inferior to supervised ones since they do not have ground truth for supervision.
	
	For supervised methods, each training data is a pair consisting of an input object (usually a vector) and an output object (ground truth). The objective is to learn a function from training data pairs to predict new examples. During training, the labels are used as supervision. For example, the function can be optimized to minimize the average difference between function outputs and labels. In \cite{zhou2008semi}, a support vector machine (SVM) was trained to extract tumor region from each 2D slice, then the extracted tumor contour was projected to its neighboring slices after some morphological steps. \cite{smeets2010semi} took a semi-automatic level set segmentation approach to segment liver tumors. Initialization is done by a dynamic programming based spiral-scanning technique; wherein the level set evolves according to a speed image generated from supervised statistical pixel classification. \cite{zhang2011interactive} used support vector machine (SVM) to detect tumors from liver parenchyma. Next, we discuss one class of supervised methods---deep convolutional neural network (DCNN).
	
	DCNN models have been widely used for images segmentation. \cite{Shelhamer2017Fully} adapted the classification networks (Alexnet, GoogLeNet and VGG net) to fully convolutional neural network (FCN) to conduct end-to-end, pixels-to-pixels semantic segmentation. Similar to FCN, in \cite{Badrinarayanan2015SegNet} an encoder-decoder architecture neural network was proposed and achieved state-of-art segmentation results even without post-processing steps. As an extension of the above studies, \cite{ronneberger2015u} used a U-shaped architecture neural network which consists of a contracting path to capture context and a symmetric expanding path that enables precise localization. They use short and long skip connections between contracting and expanding layers to recover spatial information lost during downsampling. These networks have also been used for medical images segmentation tasks. \cite{oliveira2018retinal} combines Stationary Wavelet Transform with Fully Convolutional Neural Network Network for Retinal Vessel Segmentation. \cite{havaei2017brain} used a 2-pathway cascading architecture for brain tumor segmentation. \cite{li2017h} proposed a hybrid densely connected U-Net (H-DenseUNet), which uses a 2D Dense UNet to extract intra-slice features and a 3D counterpart to aggregate volumetric contexts under the spirit of the auto-context algorithm. \cite{liu20173d} proposed a 3D Anisotropic Hybrid Network (AH-Net) that transfers convolutional features learned from 2D images to 3D anisotropic volumes. 
	
	\subsection{Motivation and contribution}
	Compared with natural images, medical images segmentation is usually more difficult. The reasons are: first, the target objects in medical images are always more irregular in size, shape and intensity, which makes it hard for the model to learn the patterns well; second, the problem of false positive is serious for medical image segmentation because the target tissues could have features (intensity, size and shape) similar to those of some other tissues; third, since annotating medical images is expensive and time-consuming, positive and negative cases in training datasets are always highly imbalanced, which can exacerbate the false positice problem. Focusing on above issues, we made the following contributions.
	
	\begin{itemize}
		\item We propose a variation of the original U-Net (Figure 2) by incorporating dense modules, inception modules and dilated convolution into the encoding path. This network is called the \textbf{``base U-Net''}. 
		
		\item We propose a \textbf{bottleneck supervised (BS) U-Net} (Figure 3) which includes an encoding U-Net and a segmentation U-Net. The encoding U-Net is first trained as an auto-encoder to learn encodings of label maps. Then, the encodings that contain anatomical information are used to guide the training of the segmentation U-Net. 
		
		\item Because the segmentation results are always worse at liver borders, we design a weight map to force the loss function to focus more on the border areas. 
		
		
		
		
	\end{itemize}
	
	\section{Technical background}

	\subsection{Segmentation network: U-Net}
	U-Net is a U-shaped convolutional neural network used for image segmentation. Figure 1 is the original U-Net architecture (cited from \cite{ronneberger2015u}). It consists of a contracting path which encodes an intensity image of size $572 \times 572$ to a $30 \times 30 \times 1024$ feature vector at the bottleneck, and an expanding path which decodes the feature vector to a label map of the same size as input intensity image. As shown in Figure 1, there are skip connections between contracting path and expanding path. The feature map at LHS of each arrow is concatenated to the feature map at RHS. For example, if the feature map at LHS is of size $3 \times 256 \times 256$ and at RHS is of size $1 \times 256 \times 256$, then the concatenated feature map is of size $(3+1) \times 256 \times 256$. These skip connections help recover information loss through the encoding path. 
	
	\begin{figure}[t]
		\centering 
		\includegraphics[height=3in]{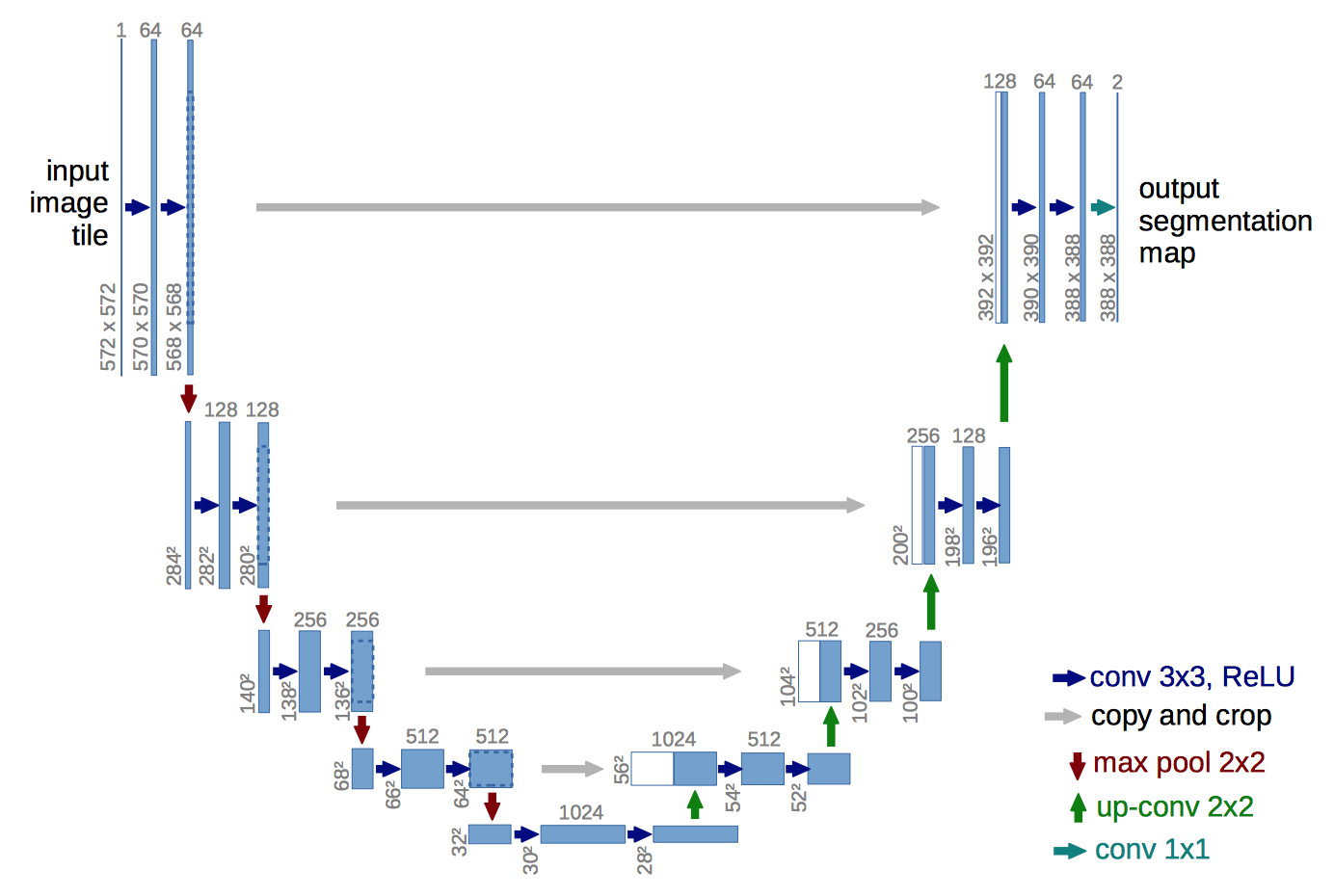}
		\caption{Original U-net architecture cited from Ronneberger et al. (2015) (example of $32 \times 32$ pixels in the lowest resolution). Each blue box corresponds to a multi-channel feature map. The number of channels is denoted on top of the box. The x-y-size is provided at the lower edge of the box. White boxes represent copied feature maps. The arrows denote the different operations.} \label{frontier}
	\end{figure}
	
	\subsection{Auto-encoder}
	An auto-encoder is a type of artificial neural network used to learn efficient data encoding in an unsupervised manner. It can be used for dimensional reduction by encoding high-dimensional data into low-dimension features. An auto-encoder usually consists of an encoder part and an decoder part. Denote the encoder function as $\phi(\cdot)$, the decoder function as $\psi(\cdot)$, an auto-encoder tries to find $ \phi, \psi = \argmin_{||\phi,\psi}(X-\phi \circ \psi(X)||^{2})$. Auto-encoder has many variants corresponding to various applications, such as denoising auto-encoder (\cite{vincent2008extracting}), sparse auto-encoder (\cite{sun2016sparse}) and variational auto-encoder (VAE) (\cite{kingma2013auto}). A feed-forward non-recurrent neural network whose input and output are the same can be deemed as an auto-encoder.
	
	\subsection{Others}
	In this paper, we additionally incorporate dense modules, inception modules and dilated convolution into the architecture of the original U-Net. For each convolution layer in the dense module, feature-maps of all preceding convolution layers are used as inputs, and its own feature-maps are used as inputs for all subsequent convolution layers. The use of dense module helps relieve the vanishing-gradient problem, strengthen feature propagation, encourage feature reuse and reduce the number of parameters. One can refer to \cite{huang2017densely} for more details. In the inception module, there are multiple paths, each with its own operation. The input is first processed independently through different paths, then the outcomes are concatenated together. There are different forms of inception module, such as the Inception V1-V4 (V1:\cite{szegedy2015going}; V2:\cite{ioffe2015batch};V3:\cite{szegedy2016rethinking};V4:\cite{szegedy2017inception}). In this paper, modules with similar multi-path structures are called `inception module' for convenience. Dilated convolution (also called `atrous convolution') is the standard convolution with kernel dilated by inserting 0s. Dilated convolution can remove pooling and generate pixels with larger reception fields, which is much desired for semantic segmentation. One can read \cite{yu2015multi} and \cite{chen2018deeplab} for more details.

	
	\section{Model and algorithm}
	This section provides details of our proposed model. The network structure of base U-Net and BS U-Net, the weight map and the loss function are presented.
	
	\subsection{Network structure}
	
	\subsubsection{Base U-Net}
	First, we propose a variation of the original U-Net that includes dense modules, inception modules and dilated convolution. Such U-Net is called \textbf{base U-Net} since it is the building block of BS U-Net.
	
	Based on the original U-Net (Figure 1), the base U-Net (Figure) incorporates an inception module, a dense module and dilated convolution in the encoder part to improve the overall performance. The left part of Figure 2 shows the structure of different blocks (parameters setting in Table 1). Each down block or transition block contains an inception module, each dense block contains a dense module. The inception module in the transition block are the same as in \cite{szegedy2015going}. The inception modules in the down block has three paths, each of which includes a dilated convolution. In the right part, `UpX', `DownX', `Trans' and `Dense' represent one up block, down block, transition block and dense block, respectively. The numbers in the parentheses are the parameters for the block (the meanings of the parameters are shown in Table 1). For example, `Down1 (8, 16)' denotes one down block with input channel = 8 and output channel = 16. `IC' and `OC' represent numbers of input and output channels of the network. The inception modules can improve encoding performance by capturing input information from different scales. The dense module used in this paper is a simplified verson, which only concatenates the input and the feature map after the second $3 \times 3$ convolution layer. It helps relieve the vanishing-gradient problem and strengthens feature propagation during bottom information transformation. A RELU activation layer follows each BN layer, and follows the convolution layers that are not followed by BN layer.
	
	\begin{figure}[t]
		\centering 
		\includegraphics[height=3in]{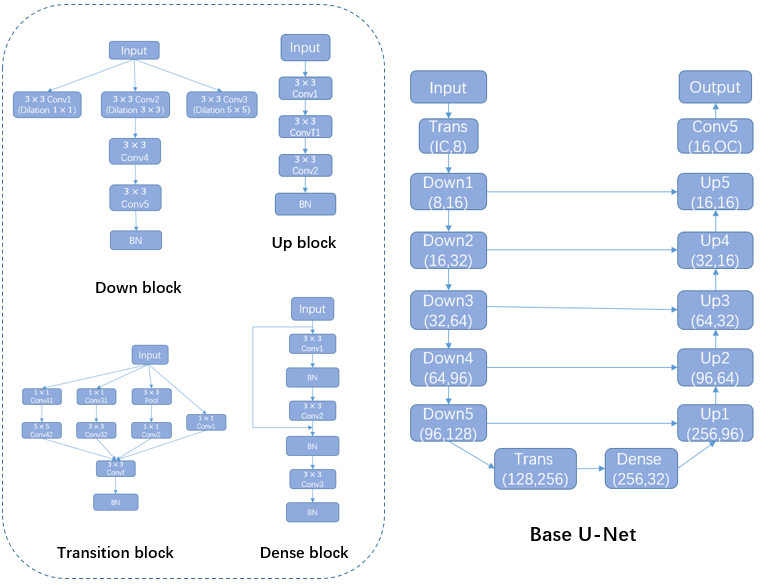}
		\caption{Base U-Net: a variation of the original U-Net.} \label{frontier}
	\end{figure}

\begin{table}[]
	\centering
	\caption{Parameters of different blocks in the base U-Net.}
	\begin{tabular}{|c|c|c|c|c|c|c|}
		\hline
                     & \textbf{In-channel}     & \textbf{Out-channel}     & \textbf{Kernel size}     & \textbf{Stride}     & \textbf{Padding}     & \textbf{Dilation}     \\ \hline
\multicolumn{7}{|c|}{\textbf{\begin{tabular}[c]{@{}c@{}}Structure of transition block \\ (input channel = a, number of output channel = b).\end{tabular}}}                \\ \hline
\textbf{Conv1}       & a                       & b                        & (1,1)                    & (1,1)               &                      &                       \\ \hline
\textbf{Pool}        & a                       & a                        & (3,3)                    & (1,1)               & (1,1)                &                       \\ \hline
\textbf{Conv2}       & a                       & b                        & (1,1)                    & (1,1)               &                      &                       \\ \hline
\textbf{Conv31}      & a                       & b                        & (1,1)                    & (1,1)               &                      &                       \\ \hline
\textbf{Conv32}      & b                       & b                        & (3,3)                    & (1,1)               & (1,1)                &                       \\ \hline
\textbf{Conv41}      & a                       & b                        & (1,1)                    & (1,1)               &                      &                       \\ \hline
\textbf{Conv42}      & b                       & b                        & (5,5)                    & (1,1)               & (2,2)                &                       \\ \hline
\textbf{Convf}       & 4b                      & b                        & (3,3)                    & (1,1)               & (1,1)                &                       \\ \hline
\textbf{BN}          &                         &                          &                          &                     &                      &                       \\ \hline
\multicolumn{7}{|c|}{\textbf{\begin{tabular}[c]{@{}c@{}}Structure of dense block \\ (input channel = a, growth rate = k).\end{tabular}}}                                  \\ \hline
\textbf{Conv1}       & a                       & k                        & (3,3)                    & (1,1)               & (1,1)                &                       \\ \hline
\textbf{BN}          &                         &                          &                          &                     &                      &                       \\ \hline
\textbf{Conv2}       & a+k                     & k                        & (3,3)                    & (1,1)               & (1,1)                &                       \\ \hline
\textbf{BN}          &                         &                          &                          &                     &                      &                       \\ \hline
\textbf{Conv3}       & a+2k                    & a                        & (3,3)                    & (1,1)               & (1,1)                &                       \\ \hline
\textbf{BN}          &                         &                          &                          &                     &                      &                       \\ \hline
\multicolumn{7}{|c|}{\textbf{\begin{tabular}[c]{@{}c@{}}Structure of up block \\ (input channel = a, inner channel = p, out channel = b, kernel size = k).\end{tabular}}} \\ \hline
\textbf{Conv1}       & a                       & p                        & (3,3)                    & (1,1)               & (1,1)                &                       \\ \hline
\textbf{ConvT1}      & p                       & b                        & (k,k)                    & (2,2)               & int(k/2,k/2)         &                       \\ \hline
\textbf{Conv2}       & b                       & b                        & (3,3)                    & (1,1)               & (1,1)                &                       \\ \hline
\textbf{BN}          &                         &                          &                          &                     &                      &                       \\ \hline
\multicolumn{7}{|c|}{\textbf{\begin{tabular}[c]{@{}c@{}}Structure of down block \\ (input channel = a, inner channels = p, output channel = b).\end{tabular}}}            \\ \hline
\textbf{Conv1}       & a                       & p                        & (3,3)                    & (1,1)               & (1,1)                & (1,1)                 \\ \hline
\textbf{Conv2}       & a                       & p                        & (3,3)                    & (1,1)               & (3,3)                & (3,3)                 \\ \hline
\textbf{Conv3}       & a                       & p                        & (3,3)                    & (1,1)               & (5,5)                & (5,5)                 \\ \hline
\textbf{Conv4}       & 3p                      & b                        & (3,3)                    & (2,2)               & (1,1)                &                       \\ \hline
\textbf{Conv5}       & b                       & b                        & (1,1)                    & (1,1)               &                      &                       \\ \hline
\textbf{BN}          &                         &                          &                          &                     &                      &                       \\ \hline

	\end{tabular}
\end{table}
		
	In practice, the dense module, the inception module and dilation convolution are found to be important for improving the liver and tumor segmentation performance. The reasons are as follows:
	\begin{itemize}
		\item \textbf{Inception module}: compared with most natural images, medical images are much more complex. For example, normal people can easily find a car in a natural image regardless of its size and surroundings. However, when looking at images that contain liver and liver tumors, only experts can find the exact borders of liver and tumors. This is because liver and liver tumors are always irregular in shape and size, and they can be similar to their surrounding organs. For example, in Figure 7 the organs around livers are quite close and similar to the livers; in Figure 8, there are many dark spots within the liver area, but only one of them is real tumor while the others are blood vessels. Similar to the experts, the inception module `looks at' the target image in different scales by using filters with different sizes. Intuitively, by doing so, we can get information of the big picture as well as details, which helps indentify the real targets and relieve the false positive issue.
		
		\item \textbf{Dense module}: dense module can significantly reduce the number of parameters and facilitate information flow. Fewer parameters make the encoding path less likely to learn from redundant information and relieves overfitting problem. In our case, less overfitting helps reduce false positives. In addition, better information flow helps relieve the `vanishing gradient' problem and reduces training time.

		\item \textbf{Dilated convolution}: each path of the inception module in the down block contains a $3 \times 3$ dilated convolution. It has been proven in the literature that dilated convolution can significantly improve segmentation results by enlarging the receptive field. What's more, our approach of including it in the incpetion module can further improve the segmentation performance by merging information with different inceptive fields.
	\end{itemize}

	One can find in figure 2 that the inception module, dense module and dilated convolution do not appear in the decoding path. That is because the decoding path only plays the role of a normal decoder and thus has no need to change. Six inception modules are included in the encoding path to better grab information from different scales at different depths. A dense module is used after encoding path to generate more precise bottleneck encoding feature vector. The idea of incorporating dense module, inception module and dilated convolution into U-Net has already been applied in the literature, such as \cite{li2017h} and \cite{dolz2018ivd}, but our model is different from theirs and achieves very excellent results.

	\subsubsection{Bottleneck supervised (BS) U-Net}
	Based on the base U-Net, we propose the bottleneck supervised (BS) U-Net (Figure 3). 
	
	First, let's look at the following fact. U-Net has an encoding path and a decoding path, and thus can be deemed as an auto-encoder if the skip connections are removed. In this paper, the above network is called the \textit{`encoding U-Net'}. After training, the feature vector at the bottleneck can be deemed as the encodings for the input. To avoid confusion, the origional U-Net used for segmentation is called the \textit{`segmentation U-Net'}. 
	
	BS U-Net is inspired by the `T-L Network' in \cite{oktay2018anatomically}. The `T-L Network' includes an auto-encoder and a predictor. The encodings generated by the well-trained auto-encoder contain anatomical information (shape, size), which is subsequently used to train the predictor by minimizing the Euclidean Loss between the encodings and the outcomes of the predictor. Similar to the `T-L' network, BS U-Net in Figure 3 includes an encoding U-Net (auto-encoder) and a segmentation U-Net (predictor). \textit{In this paper, the segmentation U-Net and encoding U-Net are set to be the base U-Net in Figure 2 with and without skip connections}. 
	
	To train the BS U-Net, the encoding U-Net is first trained with the label maps as input as well as the labels. For a well-trained encoding U-Net, the bottleneck feature vector after the dense block is the encoding for an input label map. According to \cite{oktay2018anatomically}, the encodings contain anatomical information. Subsequently, the encodings are used to train the segmentation U-Net. More specifically, the loss of the segmentation U-Net is a weighted average of the dice loss between final output and label maps and the Euclidean loss between the bottleneck feature vector of encoding U-Net and segmentation U-Net. One can see BS U-Net is similar to the `T-L network' in \cite{oktay2018anatomically}, but with encoding U-Net as the auto-encoder, segmentation U-Net as the predictor, and weighted average of Euclidean loss and dice loss as the loss function. 
	
	The approach of supervising the bottleneck feature vector comes from the following fact: \textit{given a pair of intensity image and lable map, the bottleneck feature vector of perfectly-trained encoding U-Net and segmentation U-Net should be the same}. This is correct because final outputs of both networks are the same, and therefore the bottleneck feature vector before decoding should also be the same. In this way, we add an additional supervision on the bottleneck feature vector. Our experimental results show that incorporating such information helps accelerate training, control shape distortion, and reduce false positive and false negative cases. 
	
	\begin{figure}[t]
		\centering 
		\includegraphics[height=3in]{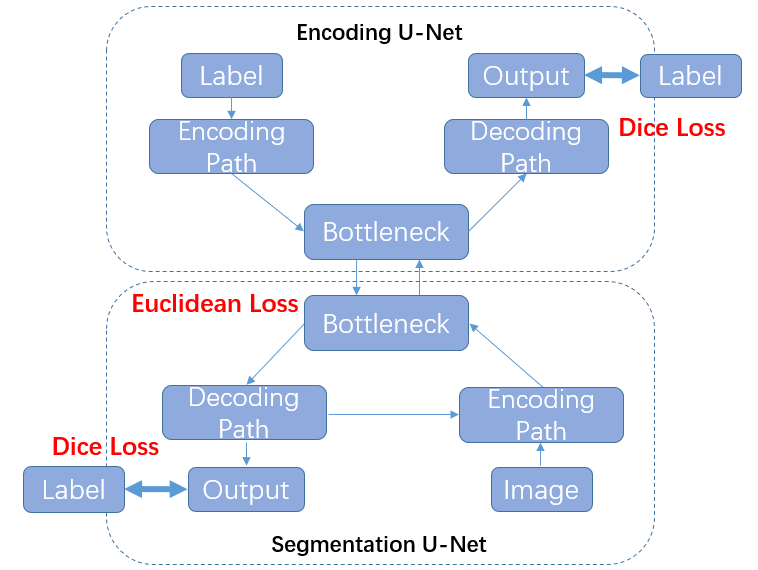}
		\caption{The bottleneck supervised (BS) U-Net} \label{frontier}
	\end{figure}
	
	\subsection{Computational complexity}
	The number of trainable parameters of the base U-Net is $6,588,139$, and of the original U-Net is $9,854,434$. Therefore, the proposed base U-Net takes less memory and is easier to train. 
	
	Compared with training a normal U-Net, training a BS U-Net requires training of an additional U-Net-like autoencoder. Actually, training the autoencoder can be very fast. In the case of liver segmentation, the dice coefficient during training of the encoding U-Net can go above 0.98 within 500 iterations. Considering the improvement of the segmentation results, such extra computational burden is affordable.
	
	\subsection{Weight function}
	When doing liver segmentation, we found that segmentation results at liver borders are always worse than results at inner areas, especially when there are tumors at the border. The reason is that the tumors, liver borders and adjacent areas can have similar intensities (relatively low). To make the neural network focus more on these `vague' areas, we compute weight maps of the same size as input intensity images and incorporate them into the loss function. Given an intensity image and its label map, a weight map is computed in two steps: first compute a distance map $D$, each pixel of which is the distance from this pixel to the nearest pixel on the liver contour; then compute the weight map by:
	\begin{equation}
	A = (w \times F + 1)e^{-\frac{D}{2 \sigma^{2}}}
	\end{equation}
	\begin{equation}
	W = \frac{A- \min{A}}{\max{A}-\min{A}}
	\end{equation}
	where $F$ is a predetermined binary (0-1) matrix called the region-of-interest (ROI) matrix. The pixels of $F$ that are equal to 1 are the interested regions. $w$ indicates the level of importance of interested regions indicated by $F$, and $\sigma$ is the variance. In practice, $\sigma$ and $w$ are selected by trial and error. Figure 4 visualizes $W$ by multiplying it by $255$. The larger $\sigma$ is, the longer is the radiation length starting from the contour. A larger weight $w$ indicates more focus on the regions induced by $F$.
	
	\begin{center}
		\begin{tabular}{ccc}	
			\includegraphics[width=0.3\linewidth]{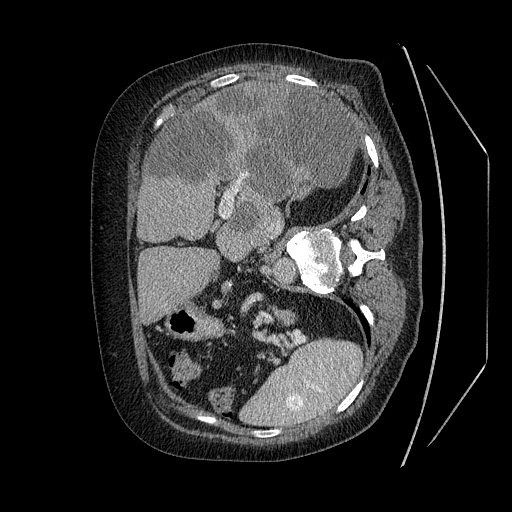} & 
			\includegraphics[width=0.3\linewidth]{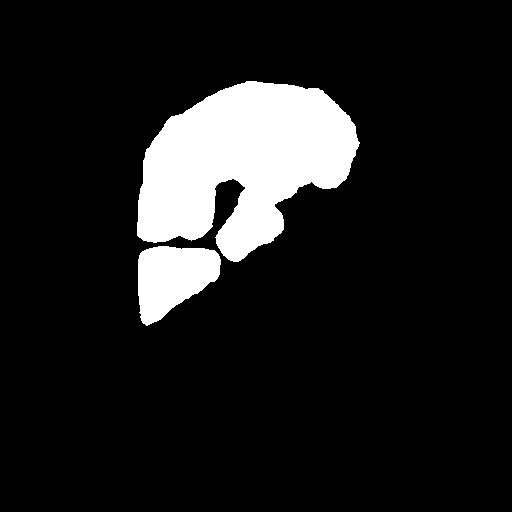} &
			\includegraphics[width=0.3\linewidth]{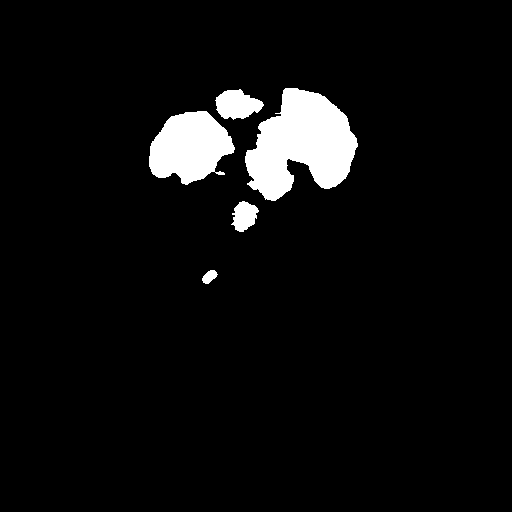} \\
			
			\includegraphics[width=0.3\linewidth]{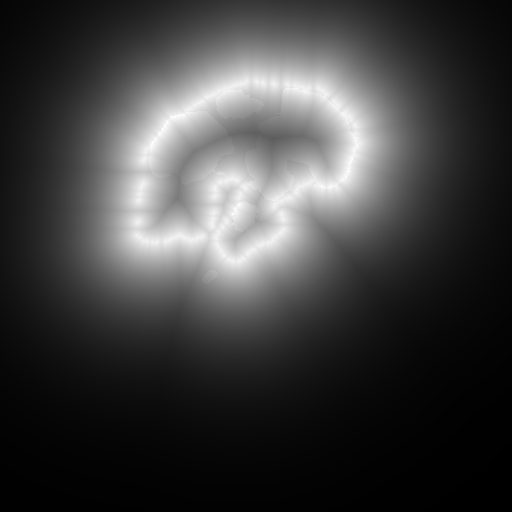} & 
			\includegraphics[width=0.3\linewidth]{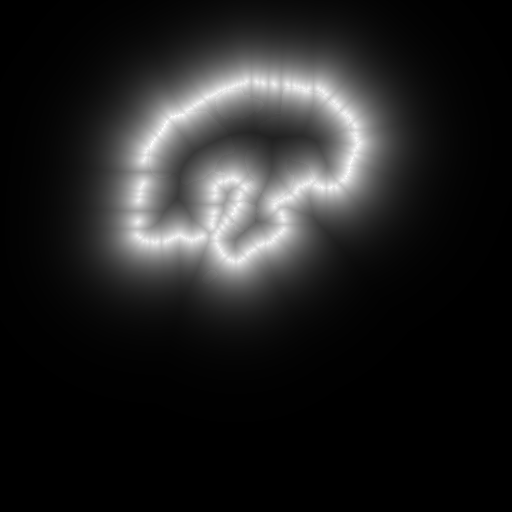} &
			\includegraphics[width=0.3\linewidth]{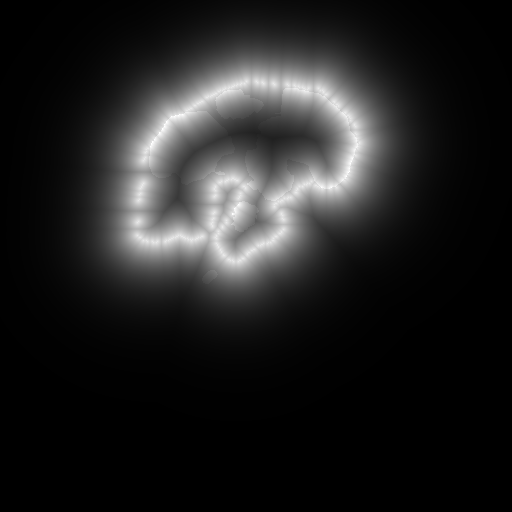} \\
			
		\end{tabular}
		\captionof{figure}{Plots of an example of original CT slice image (after preprocessing), liver segmentation image, lesion segmentation image, weight map with $\sigma = 25, w = 0.05$,  weight map with $\sigma = 15, w = 0.05$,  weight map with $\sigma = 15, w = 0.1$. }
	\end{center}
	
	\subsection{Loss function}
	To train the BS U-Net, one should first train the encoding U-Net by minimizing the dice loss between the output and the label map. Then train the segmentation U-Net by minimizing the weighted average of the weighted dice loss between the output and the label map and the Euclidean loss between the bottleneck feature vectors of the segmentation U-Net and the well-trained encoding U-Net.
	
    Given label map $A$ and output $B$, the dice loss between $A$ and $B$ is defined as:
	\begin{equation}
	Dice~loss = 1 - \frac{2 |A \bigcap B|}{|A| + |B|}
	\end{equation}
	where $|\cdot|$ computes the amount of non-zero elements. $(A\bigcap B)_{ij}=1$ if $A_{ij} = B_{ij}$ and $0$ otherwise. Dice loss has been widely used in the literature on medical image segmentation because it helps relieve data imbalance problem (\cite{milletari2016v}; \cite{sudre2017generalised}). 
	
	Denote the bottleneck feature vectors generated by encoding U-Net and segmentation U-Net as $T^{1}$ and $T^{2}$ respectively. The Euclidean loss between $T^{1}$ and $T^{2}$ is:
	\begin{equation}
	Euclidean~Loss = \sum_{i}(\bar{T}^{1}_{i}-\bar{T}^{2}_{i})^{2}
	\end{equation}
	where $\bar{T}^{1}$ and $\bar{T}^{2}$ are flattened $T^{1}$ and $T^{2}$ with one dimension. Suppose $\bar{T}^{1}$ and $\bar{T}^{2}$ are of size $n$, then $i$ goes from $1$ to $n$.
	
	To strengthen supervision on liver borders and regions induced by $F$, the weight maps generated by Equations (1) and (2) are incorporated into the dice loss. The weighted dice loss between $A$ and $B$ is computed by:
	\begin{equation}
	Dice~loss = 1 - \frac{2|W \times A \bigcap B|}{(|W \times A| + |W \times B|)}
	\end{equation}
	The total loss of BS U-Net is the weighted average of the Euclidean loss (equation (4)) amd the weighted dice loss (equation (5)).
	\begin{equation}
	Total~loss = w_{1} \times Dice~loss + w_{2} \times Euclidean~Loss
	\end{equation}
	where $w_{1} + w_{2} = 1$.

	\section{Application}
	\subsection{Dataset}
	In this section, we apply the proposed method to the public dataset of Liver Tumor Segmentation (LiTS) challenge organized by CodaLab. The official challenge website is https://competitions.codalab.org/competitions/17094. 
	
	This challenge includes liver segmentation and liver tumor segmentation. The dataset consists of 131 training datasets and 70 testing datasets, all of which are 3D abdominal CT scan images that contain liver. Label maps (annotations) are provided for the training data, but not for the testing data. Participants are required to train their model on the training data, then use the well-trained model to make predictions for the testing data. The prediction can be submitted back to the orgnizer for blind evaluation. 
	An example 3D CT data visualized in different directions is shown in Figure 5. 
	
	The CT raw data are three-dimensional matrices, elements of which are Hounsfield Units (HU). These Hounsfield Units (HU) need to be transformed into pixel intensities before feeding into the network. The preprocessing procedures are: 
	\begin{enumerate}
		\item Set HUs that are larger than 250 to be 250, lower than -200 to be -200;
		\item Normalize the outcome afte step 1 to 0-255 using MINMAX method. 
	\end{enumerate}

	\begin{figure}[h]
	\centering 
	\includegraphics[height=3.12in]{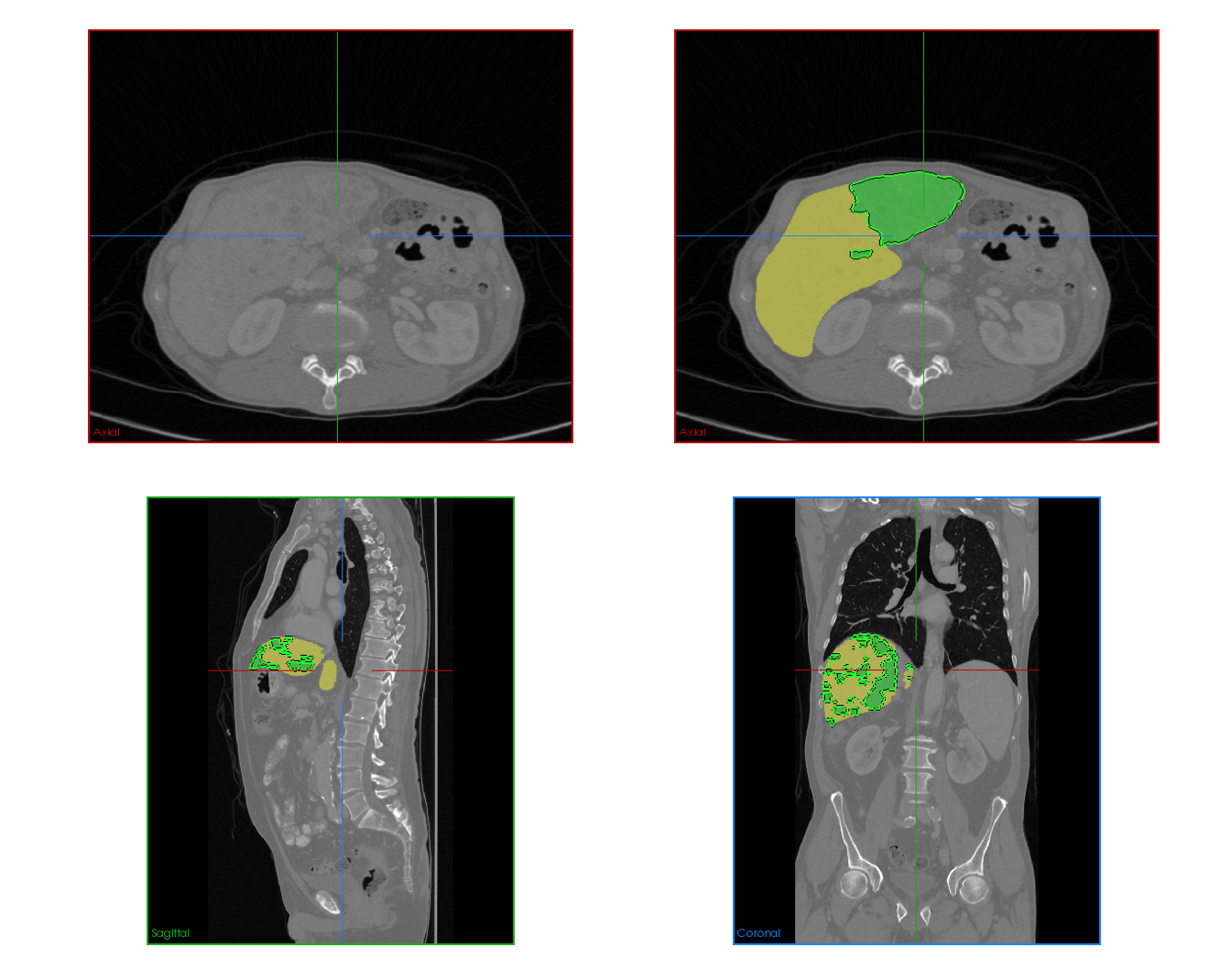}
	\caption{Different positions of an example 3D CT data. Upper left: oblique coronal position. Upper right: add annotation (label) to the upper left image, where yellow area is the liver, green areas are the tumors on the liver. Lower left: sagittalia position; lower right: oblique-axial position.} \label{frontier}
\end{figure}
	
	\subsection{Liver segmentation}
	First, we transform 3D CT data (after preprocessing) into images that can be fed to the network. We follow the routine in the literature by decomposing 3D data into 2D images along the coronal position (upper right in Figure 5). In the literature, there is a 1-channel approach and a 3-channel approach. The 1-channel approach simply decomposes 3D data into 1-channel 2D images along one dimension. For example, one $512 \times 512 \times 100$ data can be decomposed into $100$ $512 \times 512$ 1-channel images. The 3-channel approach further concatenates all three consecutive 1-channel images, thus generating $98$ $512 \times 512 \times 3$ 3-channel images. The purpose of the 3-channel approach is to introduce more contextual information.
	 
	As illustrated in Section 3.1, the training of BS U-Net includes two steps: first train the encoding U-Net, then train the segmentation U-Net. To discard redundant information, images that do not contain liver are removed from the training data. During training, the following data augmentation techniques are used: first scale all the input images to $a \times a$, where $a$ is a random number between $512$ and $600$; then randomly crop the image to $512 \times 512$. For both encoding U-Net and segmentation U-Net, the data are normalized to $[0,1]$ before putting into the network. We have conducted parameters tuning at a moderate level, the current setting is the best among all we have tried. The training parameters are set to: $batch~size = 10, initial~learning~rate = 1e-4, number~of~epochs = 50$, $ learning~rate = initial~learning~rate \times 0.3 ^{\lfloor n/3 \rfloor}$,where $n$ is the number of the current epoch. For the weight map and loss function, we let $w=0.05, \sigma=20$ in Equation (1), let $w_{1}=w_{2}=0.5$ in Equation (6). 
	
	Figure 6 shows the losses as a function of number of iterations for different models. One can see that BS U-Net, base U-Net and original U-Net all converge after some iterations. Besides, it takes fewer iterations for the base U-Net and BS U-Net to converge compared with the original U-Net. Figure 7 shows some prediction results. BS U-Net significantly reduced the occurrence of false positive (as in the first, second, and fourth rows) and false negative (the third row) cases. It is worth noting that even when the input image in the third row has low resolution, BS U-Net still successfully predicts the entire target liver. 
	
	\begin{center}
		\begin{tabular}{cc}
			\includegraphics[width=0.45\linewidth]{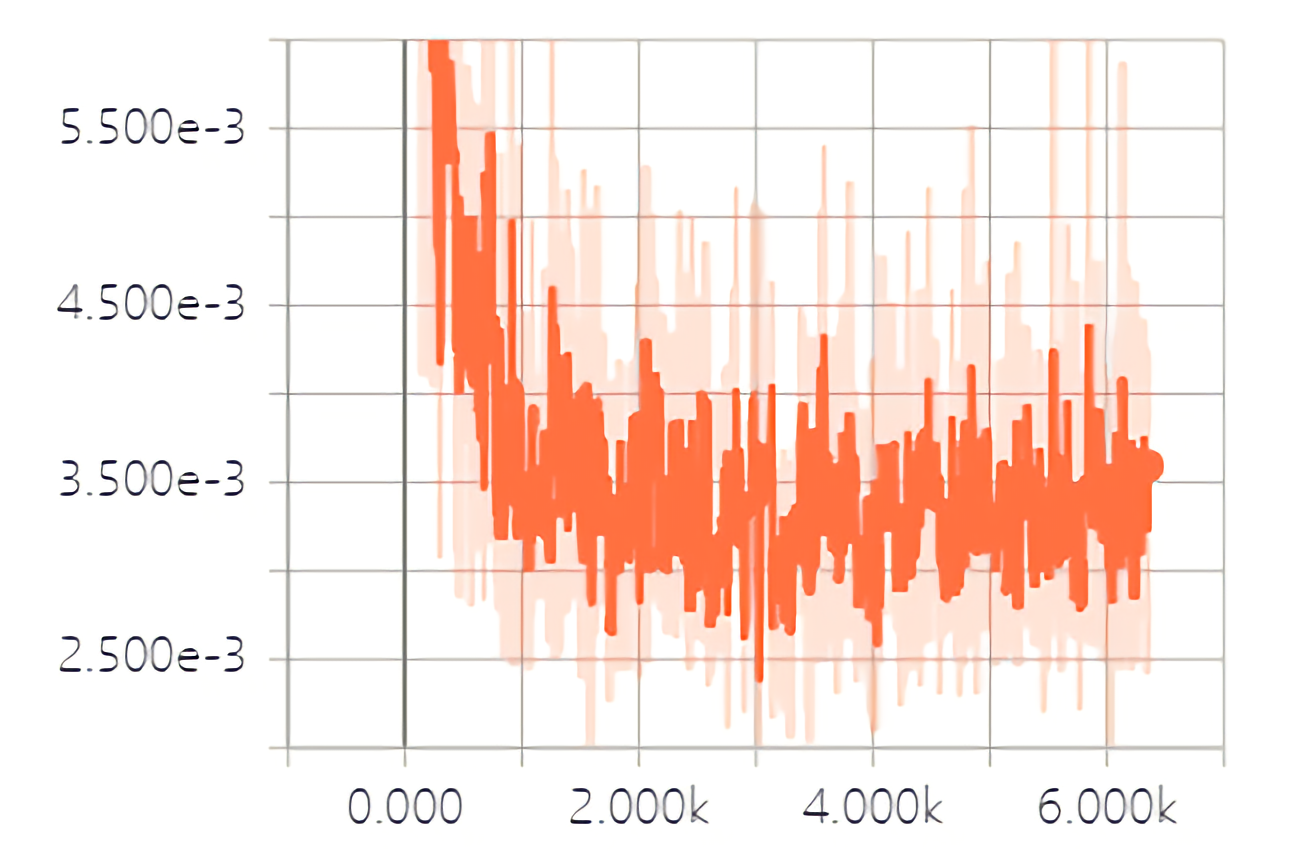} & 
			\includegraphics[width=0.45\linewidth]{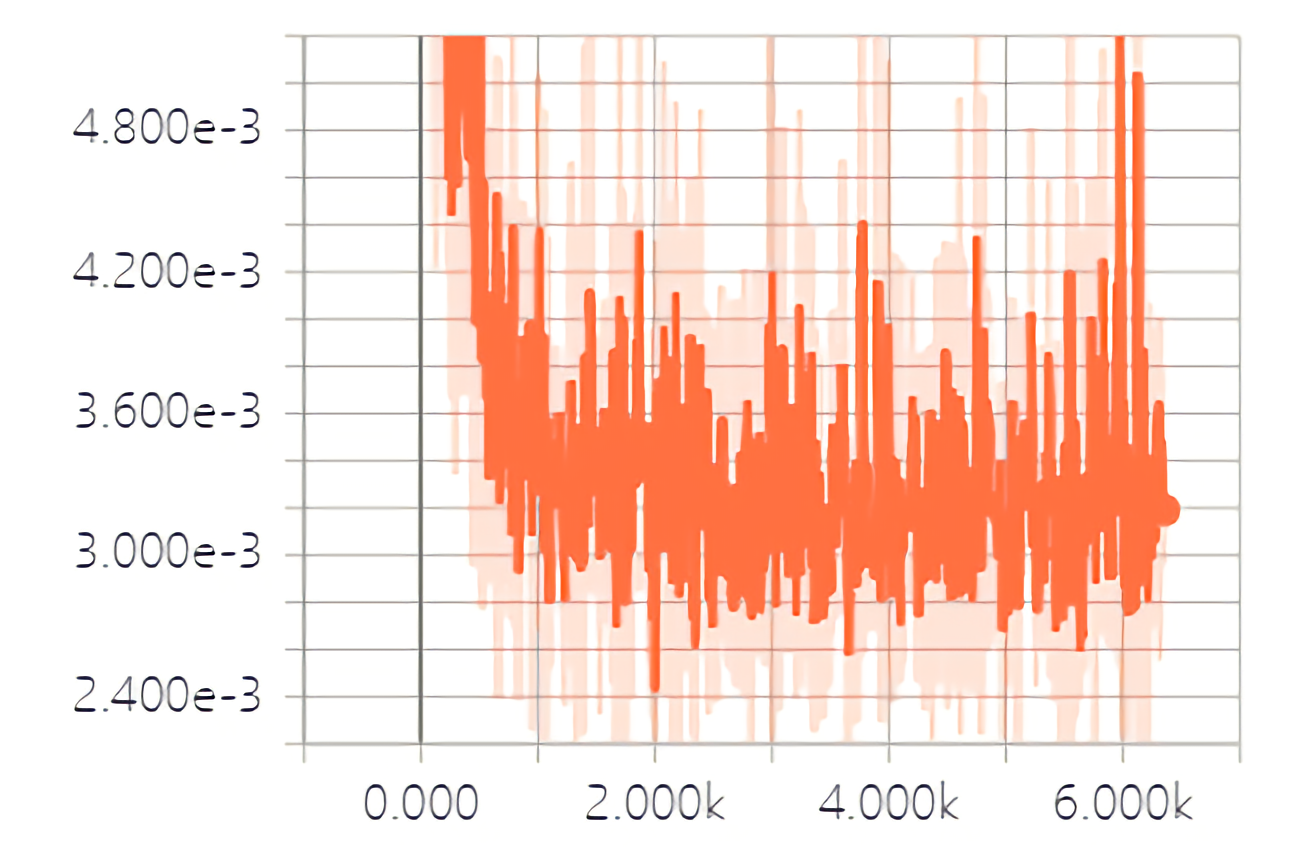} \\
			(1) & (2) \\
			\includegraphics[width=0.45\linewidth]{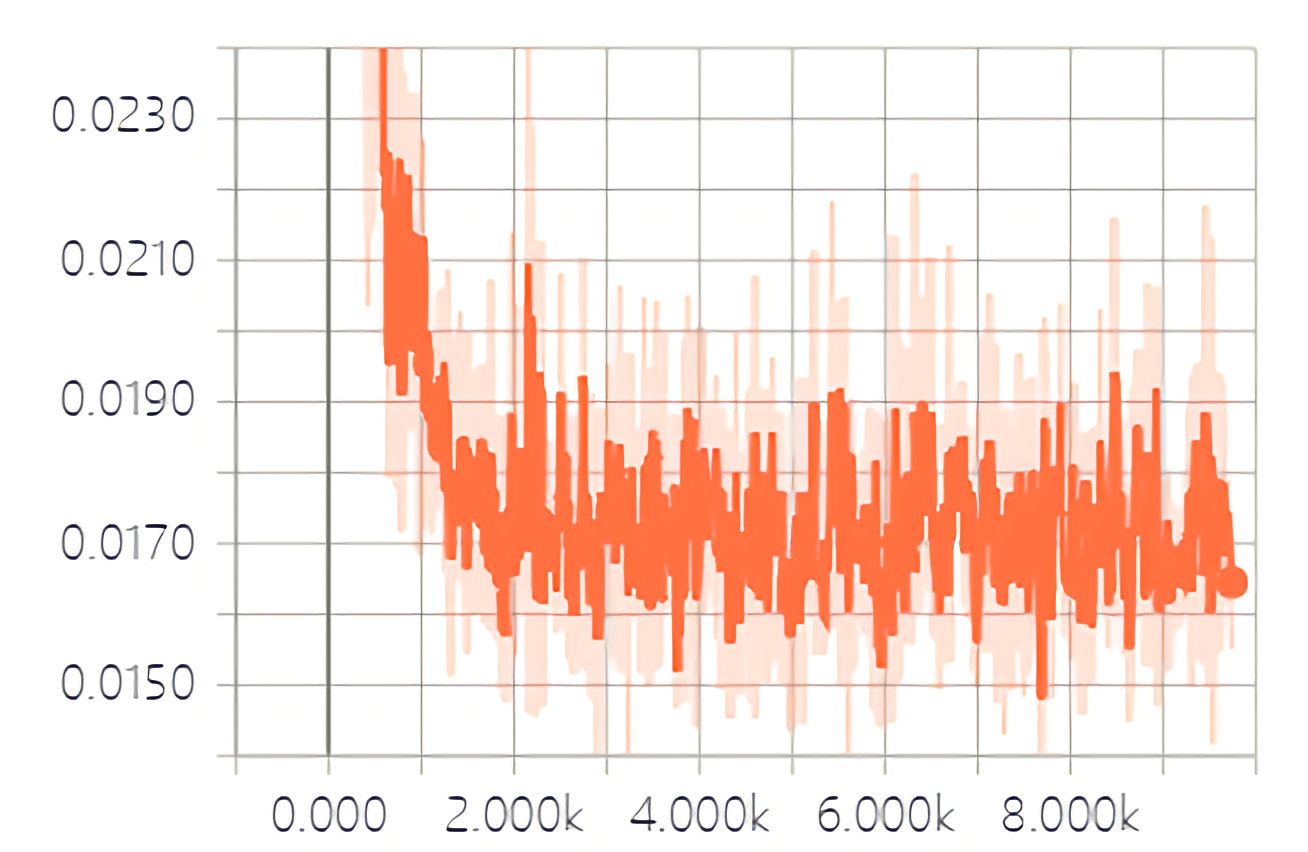} &
			\includegraphics[width=0.45\linewidth]{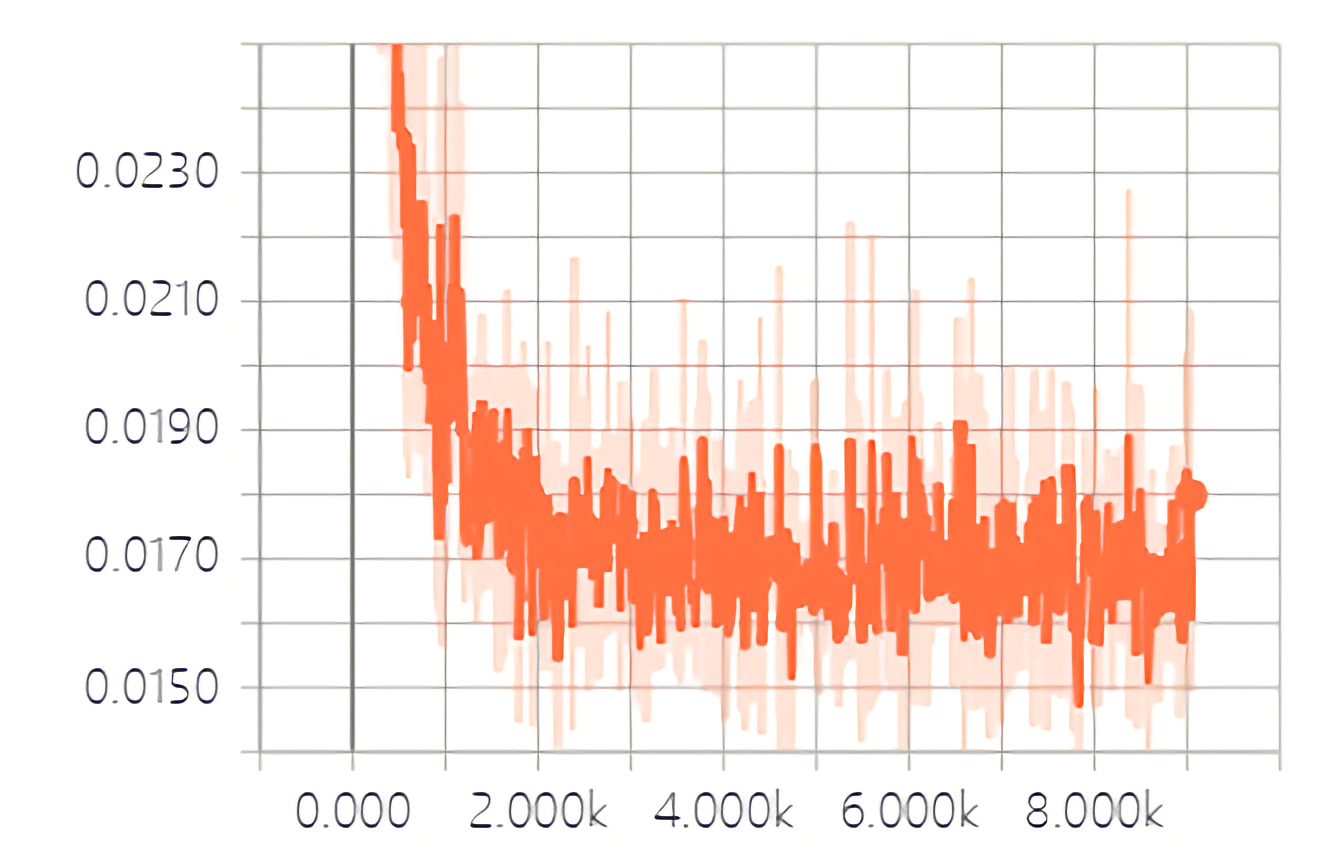}  \\
			(3) & (4) \\
			\includegraphics[width=0.45\linewidth]{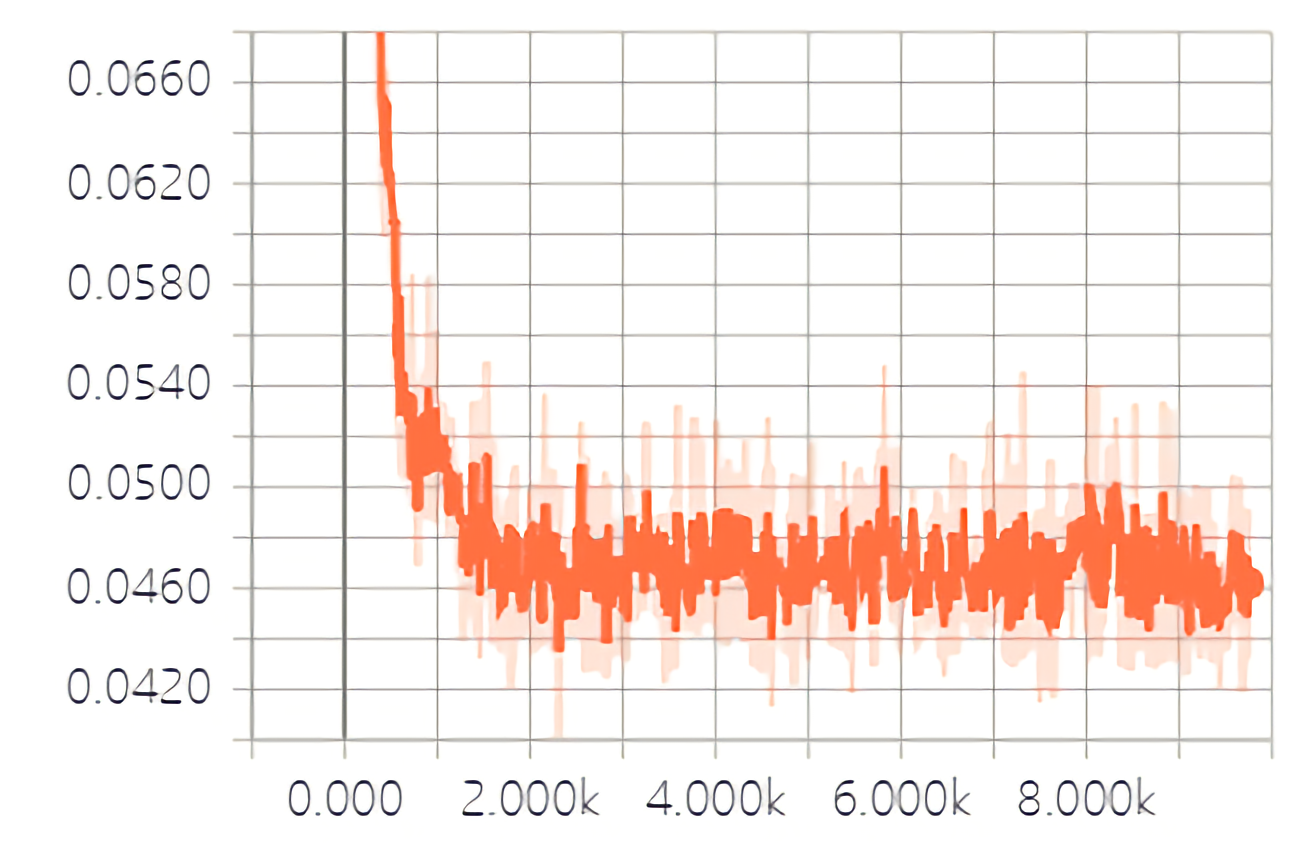} &
			\includegraphics[width=0.45\linewidth]{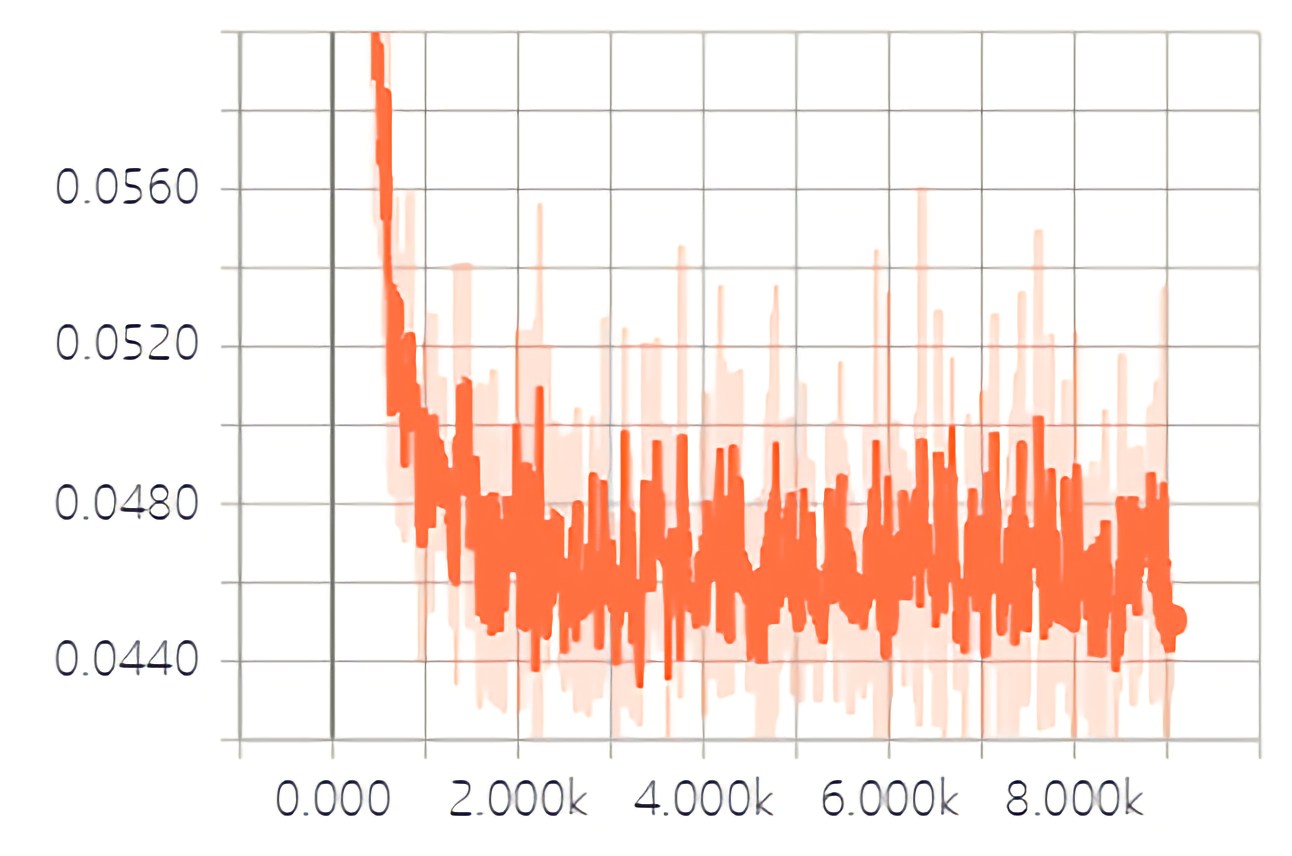} \\
			(5) & (6) \\
			\includegraphics[width=0.45\linewidth]{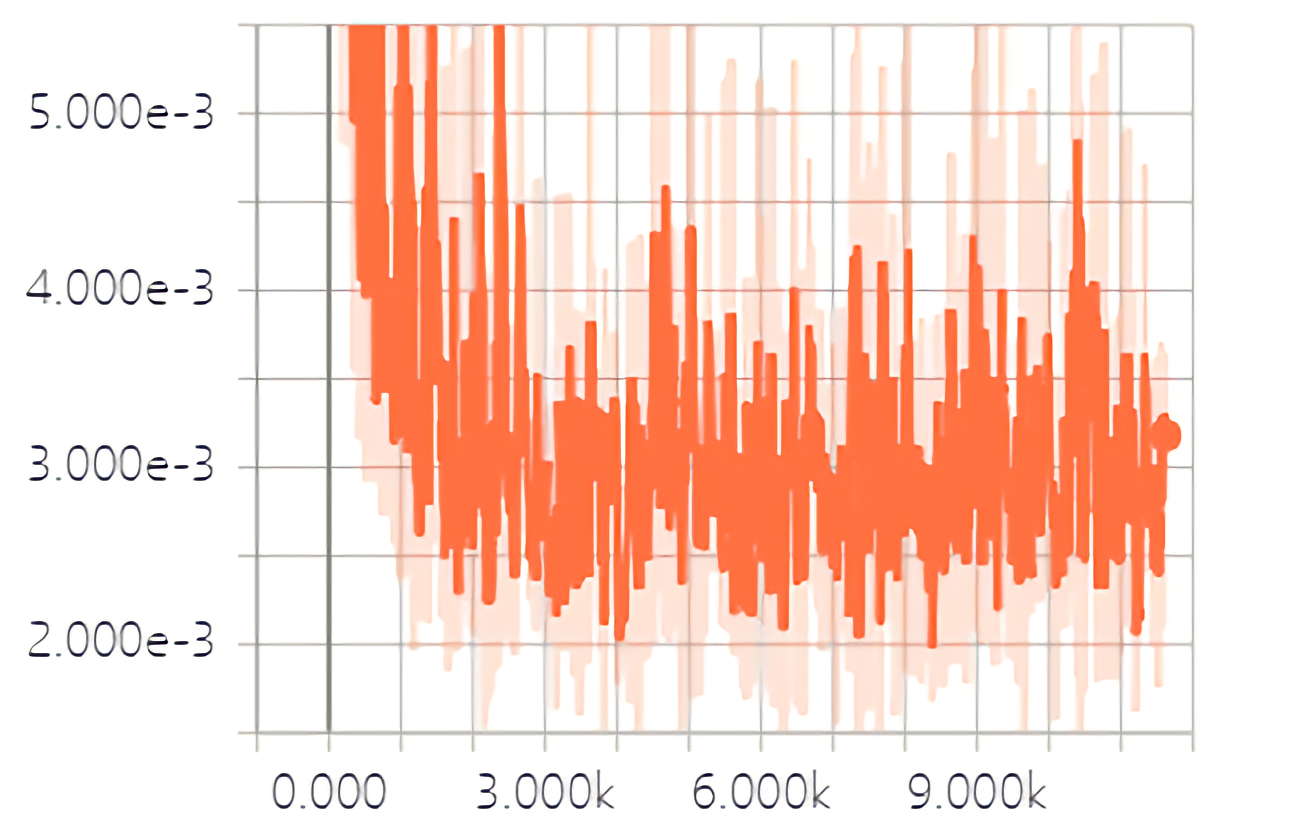} & \includegraphics[width=0.45\linewidth]{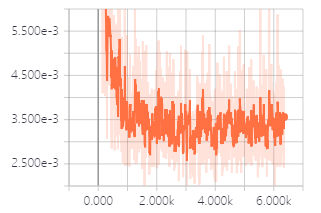}\\
			(7) & \\						
		\end{tabular}
		\captionof{figure}{Above figures are generated by `tensorboardX'. For all figures above, the horizontal axise is the number of iterations, the vertical axise is the loss. In each figure, the blurry line is the real loss, the solid line is the smoothed loss. (1) dice loss of base U-Net with 1-channel input; (2) dice loss of base U-Net with 3-channel input; (3) dice loss of BS U-Net with 1-channel input; (4) dice loss of BS U-Net with 3-channel input;  (5) Euclidean loss of BS U-Net with 1-channel input; (6) Euclidean loss of base U-Net with 3-channel input; (7) dice loss of original U-Net with 1-channel input;  (7) dice loss of original U-Net with 3-channel input.}
	\end{center}
	
	Segmentation results of base U-Net, BS U-Net and original U-Net are submitted back to the competition website (https://competitions.codalab.org/competitions/17094) for blind evaluation. The dice per case (DPC), dice global (DG), volume overlap error (VOE), relative volume difference (RVD) on the test dataset are shown in Table 1. `1C' means 1-channel approach, `3C' means 3-channel approach. In general, the 3-channel approach performs worse than 1-channel approach. Original U-Net (Figure 1) performs worse than base U-Net (Figure 2) and BS U-Net (Figure 3), meaning that including dense module, inception module and dilated convolution indeed improves the overall performance. For both 1-channel approach and 3-channel approach, the dice global of BS U-Net is 0.1 larger than base U-Net, and the dice per case of BS U-Net is 0.2 larger. The result of 1S BS U-Net ranks third in the leaderboard when this work was done. 

	\begin{table}[h]
		\centering
		\caption {Metrics on the test dataset.}
		\begin{tabular}{cccccccc}
			\hline
			& DPC & DG & VOE & RVD & ASSD & MSD & RSSD \\ \hline
			1C Ori U-Net & 0.957 & 0.960 & 0.098 & 0.053 & 1.872 & 61.235 & 4.412 \\ \hline
			3C Ori U-Net & 0.948 & 0.952 & 0.080 & 0.019 & 1.503 & 70.076 & 4.256 \\ \hline
			1C BS U-Net & 0.9610 & 0.9640 & 0.075 & 0.018 & 1.419 & 47.217 & 3.831  \\ \hline
			1C STD U-Net & 0.9590 & 0.9630 & 0.078 & 0.016 & 1.540 & 57.106 & 4.236  \\ \hline	
			
			3C BS U-Net & 0.9600 & 0.9620 & 0.077 & 0.021 & 1.543 & 55.696 & 4.307  \\ \hline
			3C STD U-Net & 0.9580 & 0.9610 & 0.079 & 0.020 & 1.470 & 76.054 & 4.174  \\ \hline	
			
		\end{tabular}
	\end{table}

	
	\clearpage
	\begin{center}
		\begin{tabular}{ccc}	
			\includegraphics[width=0.25\linewidth]{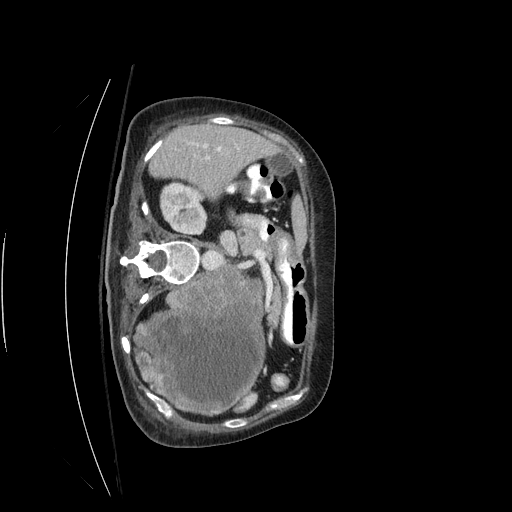} & 
			\includegraphics[width=0.25\linewidth]{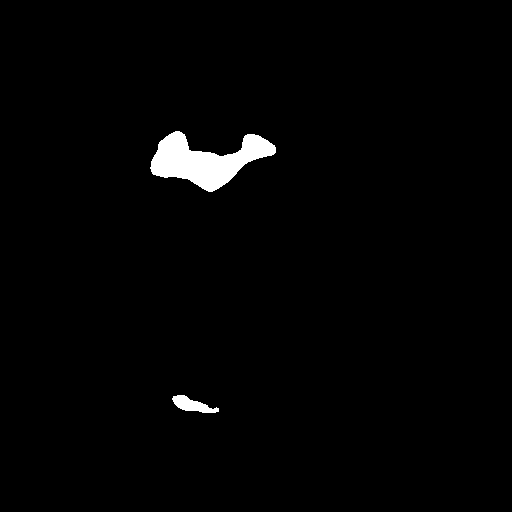} &
			\includegraphics[width=0.25\linewidth]{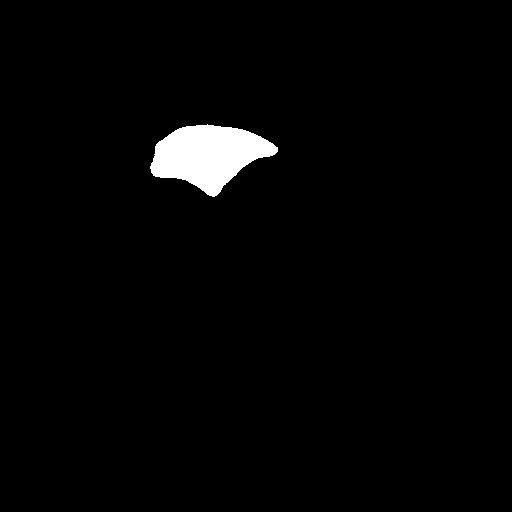} \\
			
			\includegraphics[width=0.25\linewidth]{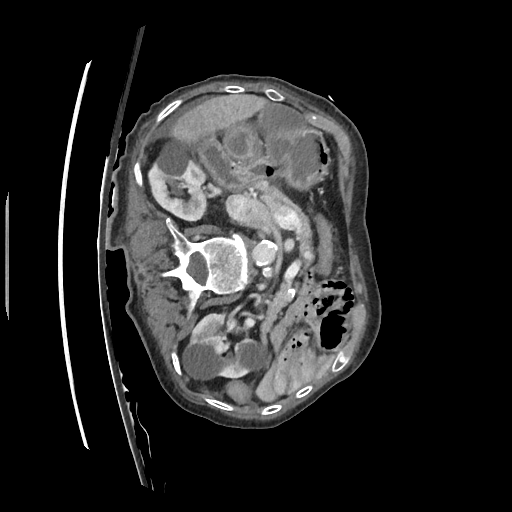} & 
			\includegraphics[width=0.25\linewidth]{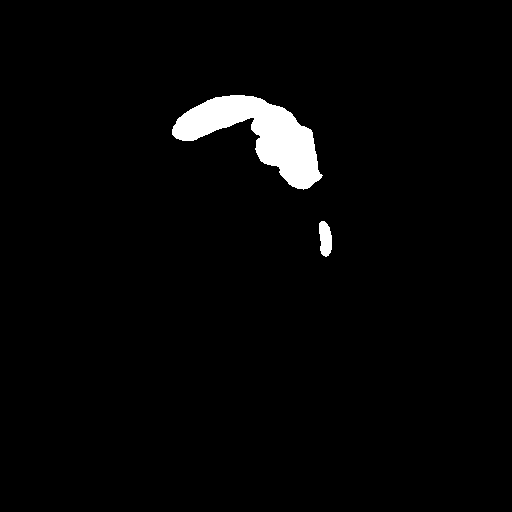} &
			\includegraphics[width=0.25\linewidth]{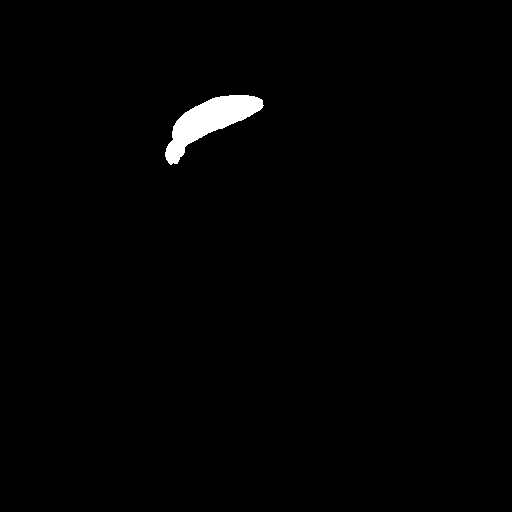} \\
			
			\includegraphics[width=0.25\linewidth]{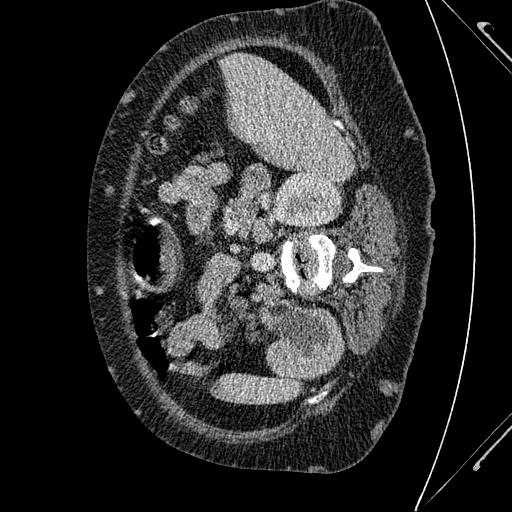} & 
			\includegraphics[width=0.25\linewidth]{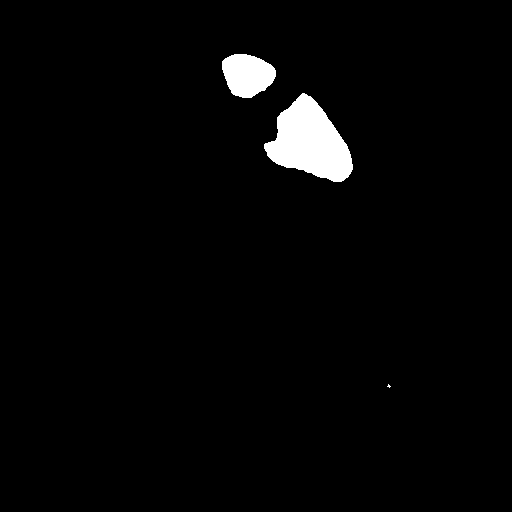} &
			\includegraphics[width=0.25\linewidth]{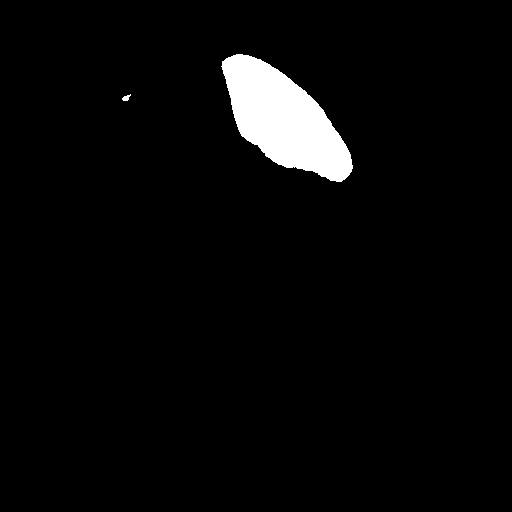}\\
			
			\includegraphics[width=0.25\linewidth]{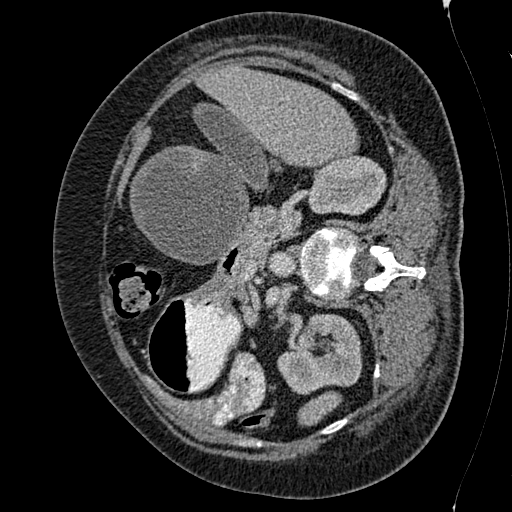} & 
			\includegraphics[width=0.25\linewidth]{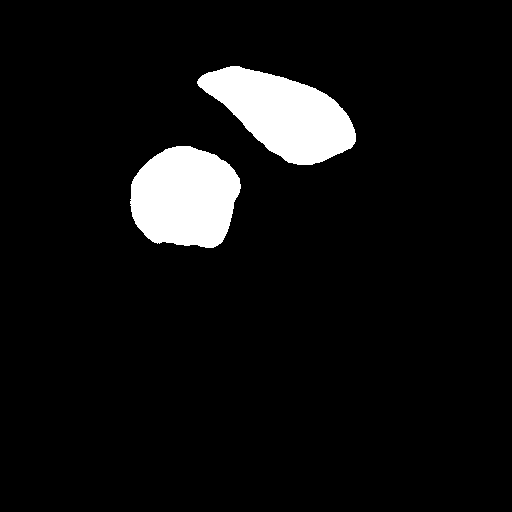} &
			\includegraphics[width=0.25\linewidth]{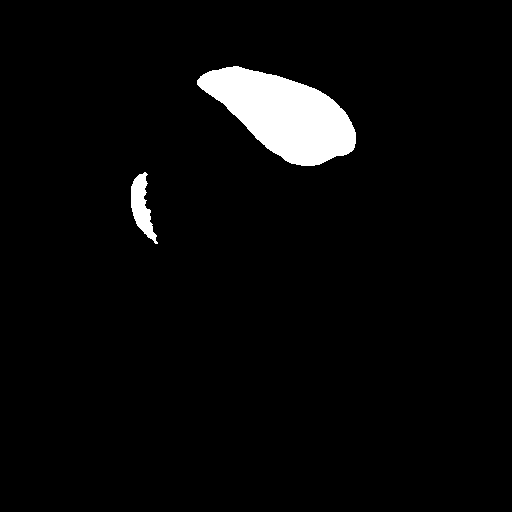}\\
			
			Input image & base U-Net & BS U-Net
		\end{tabular}
		\captionof{figure}{The first column shows the input image before normalization, the second column shows segmentation results using base U-Net model, the third column shows segmentation results using bottleneck supervised (BS) U-Net.}	
	\end{center}
	
	
	\subsection{Tumor segmentation}
	In this subsection, we use BS U-Net to conduct tumor segmentation based on the liver segmentation results in Section 4.1. The two sequential neural networks system---one for liver segmentation and one for tumor segmentation---is called the cascaded neural network structure. Same as in liver segmentation, the encoding U-Net and segmentation U-Net in BS U-Net are set to be the base U-Net in Figure 2 with and without skip connections. The hyperparameters setting is: $batch~size = 20, initial~learning~rate = 1e-4, number~of~epochs = 50$, $learning~rate = initial~learning~rate \times 0.3 ^{\lfloor n/3 \rfloor}$, where $n$ is the number of current epoch. For the loss function, we set $w_{1}=w_{2}=0.5$ in Equation (6). To discard redundant information and relieve computation burden, we preprocess the images and include only images that contain liver in the training dataset. Figure 8 shows the scheme of preprocessing :
	\begin{enumerate}
		\item Use liver mask to mask out the background; 
		\item Crop out a rectangular area containing the liver with $10$ pixels margin at the top, bottom, left and right; 
		\item Pad the cropped image to a square, whose size takes the maximum of cropped image's length and width; 
		\item Rescale the square image to $224 \times 224$.
	\end{enumerate}
	
	\begin{figure}[t]
		\centering 
		\includegraphics[height=3.12in]{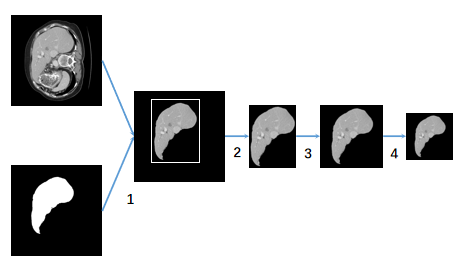}
		\caption{The pipeline of image preprocessing for tumor segmentation.} 
		\label{frontier}
	\end{figure}
	
	During training, the following data augmentation methods are used: 1. first scale the images (after preprocessing) to $300 \times 300$, then randomly crop to $224 \times 224$; 2. randomly rotate the images with angle randomly selected from $-45^{\circ}$ to $45^{\circ}$. Figure 9 shows the losses as a function of number of iterations. Segmentation results of BS U-Net and base U-Net are submitted back to the official competition website, feedback metrics are shown in Table 2. Both dice per case and dice global of BS U-Net are better than base U-Net. Meaning that the supervision on bottleneck feature vectors is important.
	
	\begin{table}
		\centering
		\caption {Metrics on the test dataset.}
		\begin{tabular}{cccccccc}
			\hline
			& DPC & DG & VOE & RVD & ASSD & MSD & RSSD \\ \hline
			BS U-Net & 0.569 & 0.751 & 0.437 & -0.228 & 1.702 & 9.130 & 2.426  \\ \hline
			STD U-Net & 0.5520 & 0.7290 & 0.414 & -0.101 & 1.395 & 8.324 & 2.069  \\ \hline	
			
		\end{tabular}
	\end{table}
	
	\begin{center}
		\begin{tabular}{ccc}	
			\includegraphics[width=0.31\linewidth]{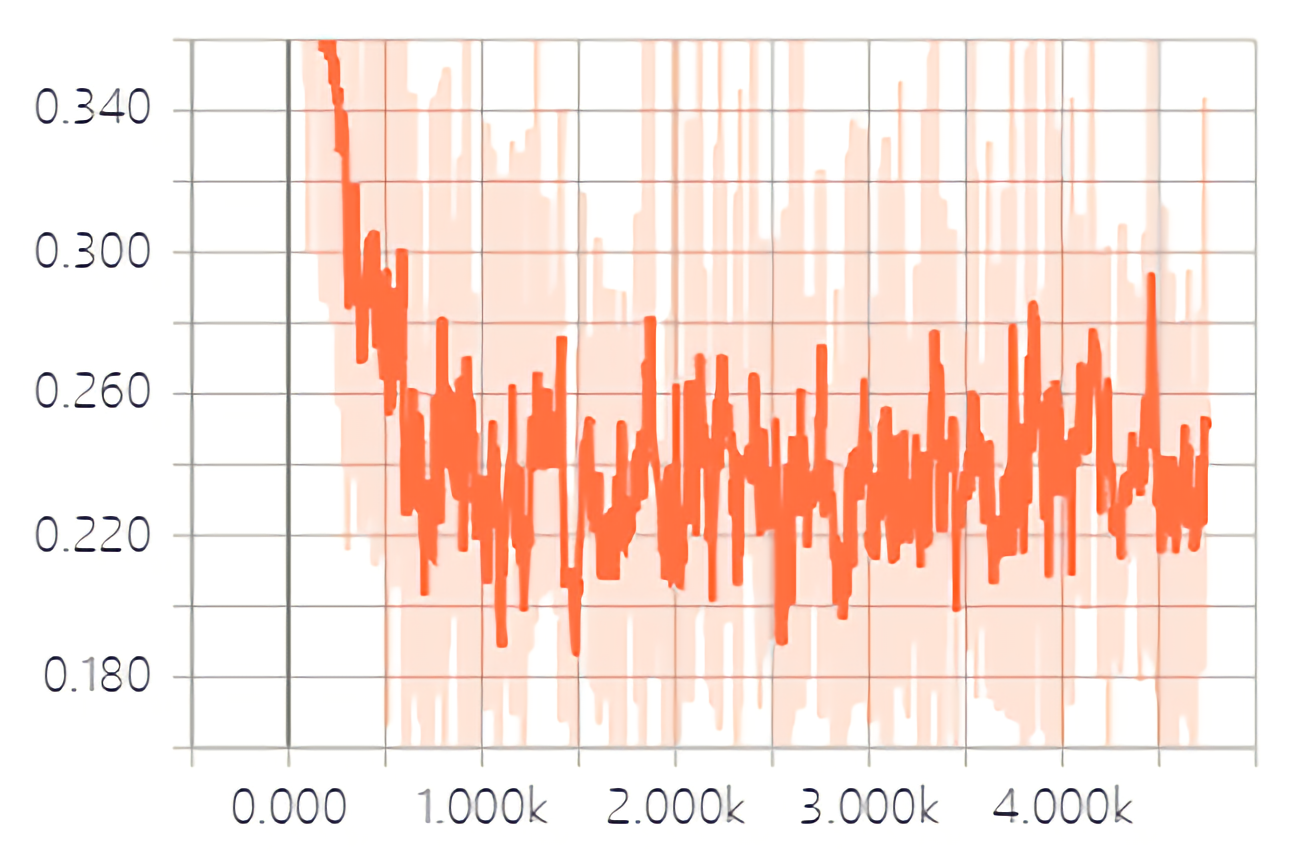} & 
			\includegraphics[width=0.31\linewidth]{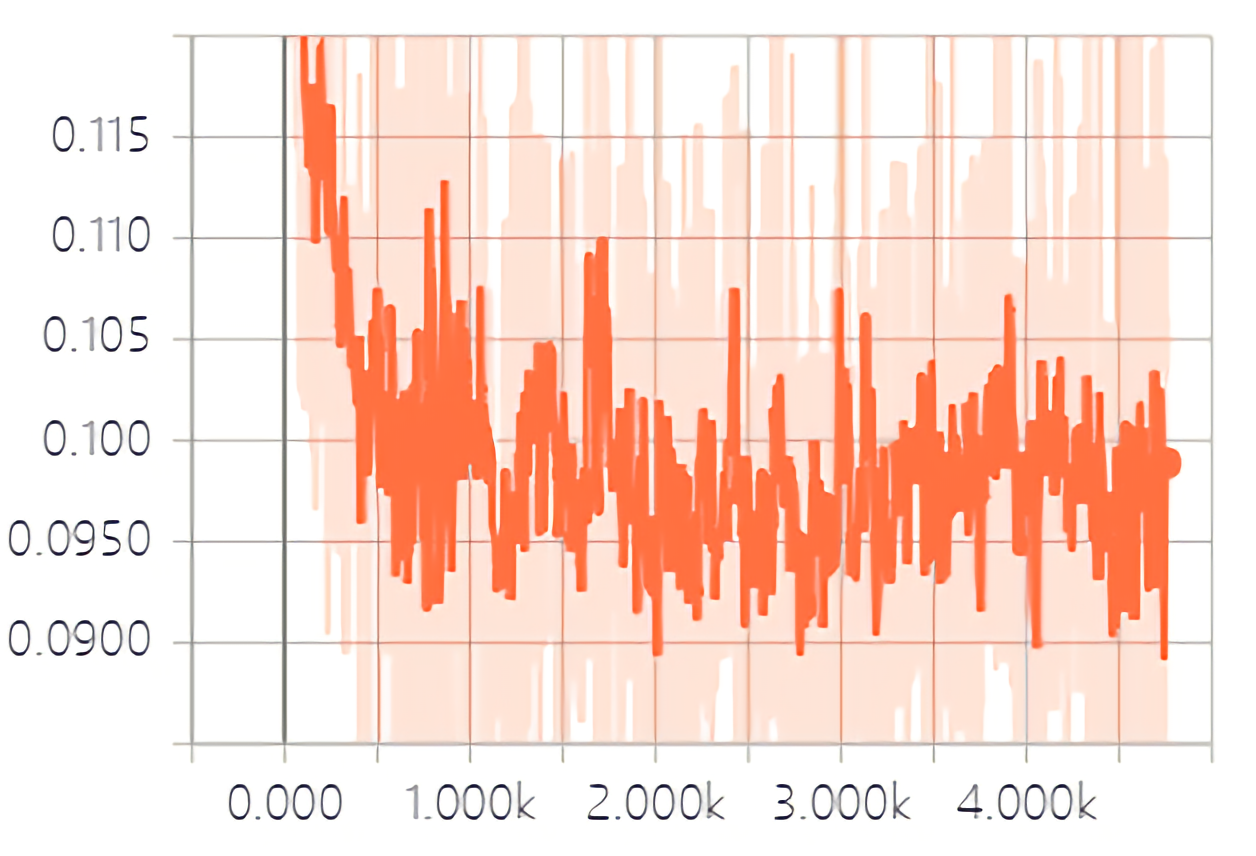}  &	
			\includegraphics[width=0.31\linewidth]{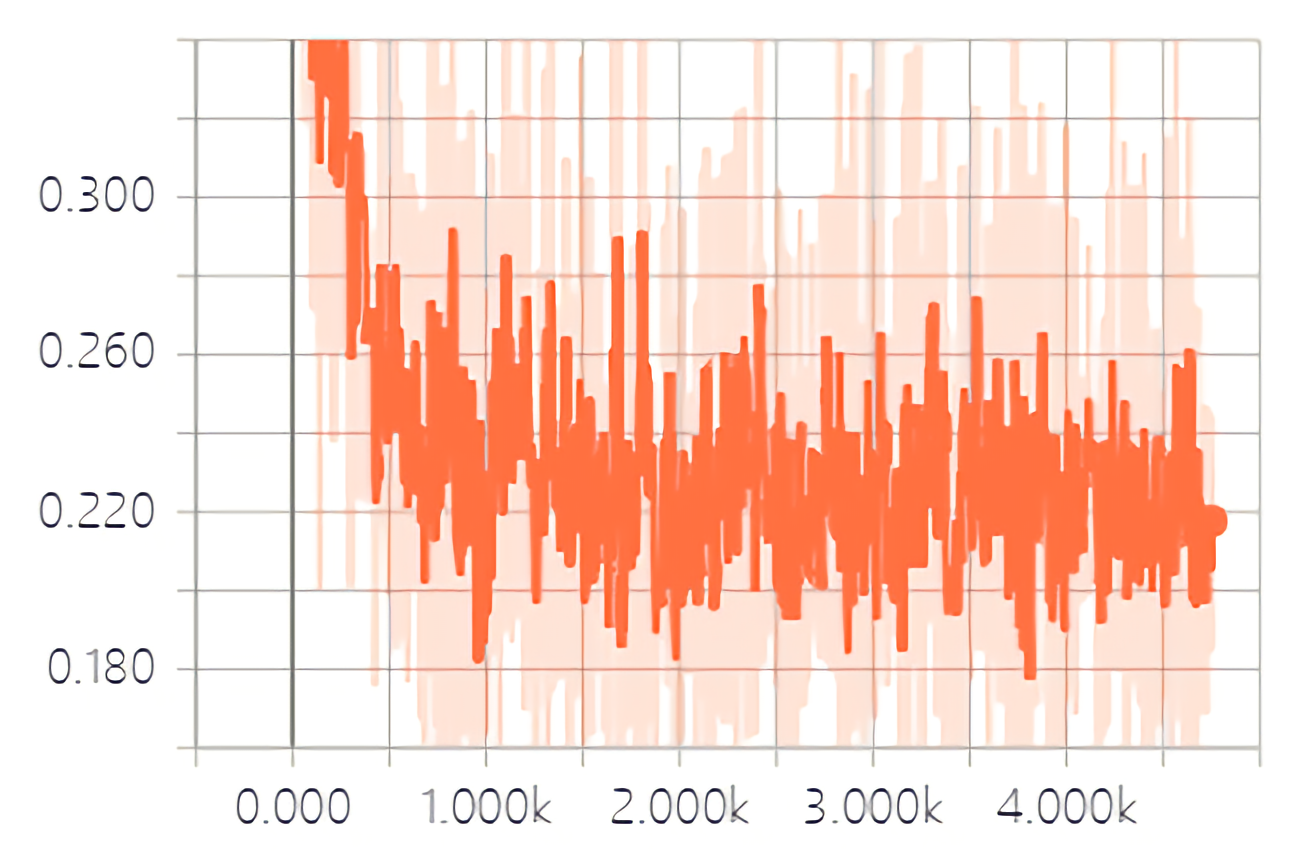} \\ 
			(1) & (2) & (3) \\
		\end{tabular}
		\captionof{figure}{Above figures are generated by `tensorboardX'. For all figures above, the horizontal axise is the number of iterations, the vertical axis is the loss. In each figure, the blurry line is the real loss, the solid line is the smoothed loss. (1) dice loss of BS U-Net; (2) MSE loss of BS U-Net; (3) dice loss of base U-Net.}	
	\end{center}
	
	\section{Conclusion and discussion}
	In this paper, we propose a novel BS U-Net to conduct liver and liver tumor segmentation. More specifically, we first propose a variation of original U-Net that includes dense modules, inception modules and dilated convolution in the encoding path. Then we propose the BS U-Net that includes an encoding U-Net and a segmentation U-Net. During training, we first train the encoding U-Net, then train the segmentation U-Net guided by the encodings generated by the well-trained encoding U-Net. After training, the model is tested on the test dataset of public LiTS dataset; and results are submitted to the competition website for blind evaluation. Feedback metrics show that the dice per case and dice global measures are largest for BS U-Net and smallest for original U-Net. In addition, prediction results (Figure 7) show that BS U-Net has performed better in controlling shape distortion, reducing false positive and false negative cases.
	
	The BS U-Net can be generalized by varying the structure of base U-Net. With a novel U-Net that has achieved good performance, one can use it as the base U-Net to construct and train the BS U-Net for further improvement. In this way, BS U-Net has very good potential. 
	

	
	\bibliography{ref}
	
\end{document}